\def\year{2018}\relax
\documentclass[letterpaper]{article} 
\usepackage{Format/aaai18}  
\usepackage{times}  
\usepackage{helvet}  
\usepackage{courier}  
\usepackage{url}  
\usepackage{graphicx}  
\usepackage[only-used=true, single=false]{acro}
\usepackage[ruled,vlined,linesnumbered]{algorithm2e}
\usepackage{amsfonts}
\usepackage{amsmath, amssymb}
\usepackage{array}
\usepackage{bm}
\usepackage{color}
\usepackage[utf8]{inputenc} 
\usepackage[T1]{fontenc}    
\usepackage{subfigure}
\usepackage{mathtools}

\nocopyright

\title{\LARGE \bf Learning and Generalisation of Primitives Skills \\ Towards Robust Dual-arm Manipulation}
\author{\`{E}ric Pairet \and Paola Ard\'{o}n \and Frank Broz \and Michael Mistry \and Yvan Petillot\\
Edinburgh Centre for Robotics\\
University of Edinburgh and Heriot-Watt University, UK\\
{\tt\small \{eric.pairet, paola.ardon\}@ed.ac.uk}, {\tt\small f.broz@hw.ac.uk}, {\tt\small mmistry@inf.ed.ac.uk}, {\tt\small y.r.petillot@hw.ac.uk}\\
}

\newcommand*{\cref}[1]{Chapter~\ref{#1}}

\newcommand*{\eref}[1]{(\ref{#1})}
\newcommand*{\fref}[1]{Figure~\ref{#1}}

\newcommand{\real}{\mathbb{R}}

\newcommand{\dimcs}{N} 

\newcommand{\workspace}{\mathcal{W}}
\newcommand{\object}{\mathcal{O}}

\newcommand{\graspmatrix}{\mathbf{G}}

\newcommand{\vf}{\mathbf{f}}
\newcommand{\vw}{\mathbf{w}}

\newcommand{\vk}{\mathbf{k}}

\newcommand{\pos}{x}
\newcommand{\vel}{\dot{x}}
\newcommand{\acc}{\ddot{x}}

\newcommand{\vx}{\mathbf{x}}
\newcommand{\vpos}{\mathbf{x}}
\newcommand{\vvel}{\mathbf{\dot{x}}}
\newcommand{\vacc}{\mathbf{\ddot{x}}}

\newcommand{\vgoal}{\mathbf{g}}
\newcommand{\Askill}{\mathcal{S}_a}
\newcommand{\Rskill}{\mathcal{S}_r}

\newcommand{\vr}{\mathbf{r}}

\newcommand{\adaptation}[1]{\mathbf{f}{#1}(\boldsymbol{\cdot})}

\DeclareAcronym{1D}{
  short = 1D,
  long  = one-dimensional
}
\DeclareAcronym{2D}{
  short = 2D,
  long  = two-dimensional
}
\DeclareAcronym{3D}{
  short = 3D,
  long  = three-dimensional
}
\DeclareAcronym{AI}{
  short = AI,
  long  = artificial intelligence
}
\DeclareAcronym{CAN}{
  short = CAN,
  long  = controller area network
}
\DeclareAcronym{CF}{
  short = CF,
  long  = coupling force
}
\DeclareAcronym{CMP}{
  short = CMP,
  long  = compliant movement primitive
}
\DeclareAcronym{DMP}{
  short = DMP,
  long  = dynamic movement primitive
}
\DeclareAcronym{DoF}{
  short = DoF,
  long  = degree of freedom,
  long-plural-form = degrees of freedom
}
\DeclareAcronym{DS}{
  short = DS,
  long  = dynamical system
}
\DeclareAcronym{ECR}{
  short = ECR,
  long  = Edinburgh Centre for Robotics
}
\DeclareAcronym{EM}{
  short = EM,
  long  = expectation-maximisation
}
\DeclareAcronym{FK}{
  short = FK,
  long  = forwad kinematics
}
\DeclareAcronym{GMM}{
  short = GMM,
  long  = Gaussian mixture model
}
\DeclareAcronym{GMR}{
  short = GMR,
  long  = Gaussian mixture regression
}
\DeclareAcronym{GPR}{
  short = GPR,
  long  = Gaussian process regression
}
\DeclareAcronym{HMM}{
  short = HMM,
  long  = hidden Markov model
}
\DeclareAcronym{HRI}{
  short = HRI,
  long  = human-robot interaction
}
\DeclareAcronym{HSMM}{
  short = HSMM,
  long  = hidden semi-Markov model
}
\DeclareAcronym{KL}{
  short = KL,
  long  = Kullback-Leibler
}
\DeclareAcronym{IGMM}{
  short = IGMM,
  long  = infinite Gaussian mixture model
}
\DeclareAcronym{IIT}{
  short = IIT,
  long  = Italian Institute of Technology
}
\DeclareAcronym{IK}{
  short = IK,
  long  = inverse kinematics
}
\DeclareAcronym{ILC}{
  short = ILC,
  long  = iterative learning control
}
\DeclareAcronym{IR}{
  short = IR,
  long  = infra-red
}
\DeclareAcronym{LbD}{
  short = LbD,
  long  = learning by demonstration
}
\DeclareAcronym{LED}{
  short = LED,
  long  = light-emitting diode
}
\DeclareAcronym{LMS}{
  short = LMS,
  long  = least mean squares
}
\DeclareAcronym{LS}{
  short = LS,
  long  = linear square
}
\DeclareAcronym{LWPR}{
  short = LWPR,
  long  = locally weighted projection regression
}
\DeclareAcronym{LWR}{
  short = LWR,
  long  = locally weighted regression
}
\DeclareAcronym{PD}{
  short = PD,
  long  = proportional-derivative
}
\DeclareAcronym{RBF}{
  short = RBF,
  long  = radial basis function
}
\DeclareAcronym{RFWR}{
  short = RFWR,
  long  = receptive field weighted regression
}
\DeclareAcronym{RL}{
  short = RL,
  long  = reinforcement learning
}
\DeclareAcronym{ROS}{
  short = ROS,
  long  = robot operating system
}
\DeclareAcronym{WP}{
  short = WP,
  long  = work package
}
\DeclareAcronym{YARP}{
  short = YARP,
  long  = yet another robotic platform
}

\begin{document}
	%
	\maketitle

	\begin{abstract}
    Robots are becoming a vital ingredient in society. Some of their daily tasks require dual-arm manipulation skills in the rapidly changing, dynamic and unpredictable real-world environments where they have to operate. Given the expertise of humans in conducting these activities, it is natural to study humans’ motions to use the resulting knowledge in robotic control. With this in mind, this work leverages human knowledge to formulate a more general, real-time, and less task-specific framework for dual-arm manipulation. The proposed framework is evaluated on the iCub humanoid robot and several synthetic experiments, by conducting a dual-arm pick-and-place task of a parcel in the presence of unexpected obstacles. Results suggest the suitability of the method towards robust and generalisable dual-arm manipulation.
\end{abstract}

	\section{INTRODUCTION} \label{sec:introduction}
    The last decades have witnessed a drastic increase in the use of robots in industry, professional and domestic environments. Among the countless competences that robots have acquired, some of the most outstanding are automating repetitive and exhausting tasks in manufacturing plants, working in hazardous scenarios unreachable to humans, assisting doctors in challenging surgical operations, and taking the responsibility for household chores. A common issue in all these applications is the need of manipulating large objects and ensembling multi-component elements without external assistance. On top of that, current manipulators lack human-like generalisation capabilities to confront the highly dynamic and changing environments. Thus, endowing robots with human-like dual-arm manipulation capabilities is essential to extend their competences and autonomy.

    Traditional approaches for governing these dual-arm systems depend upon a great understanding of the model underlying the system’s behaviour~\cite{smith2012dual}. Even though deriving an accurate model is possible for some complex systems, approximations are commonly used in order to make the calculations computationally tractable, despite the trade-off of the model’s uncertainty~\cite{pairet2018uncertainty}. Furthermore, some of these methods lack scalability and generalisation capabilities along and across tasks. In other words, they require an expert programmer to hand-define all possible scenarios, movements, tasks, and extensive manual tuning of the system’s control architecture~\cite{argall2009survey}.

    \begin{figure}[t!]
        \centering
        \includegraphics[width=5.5cm]{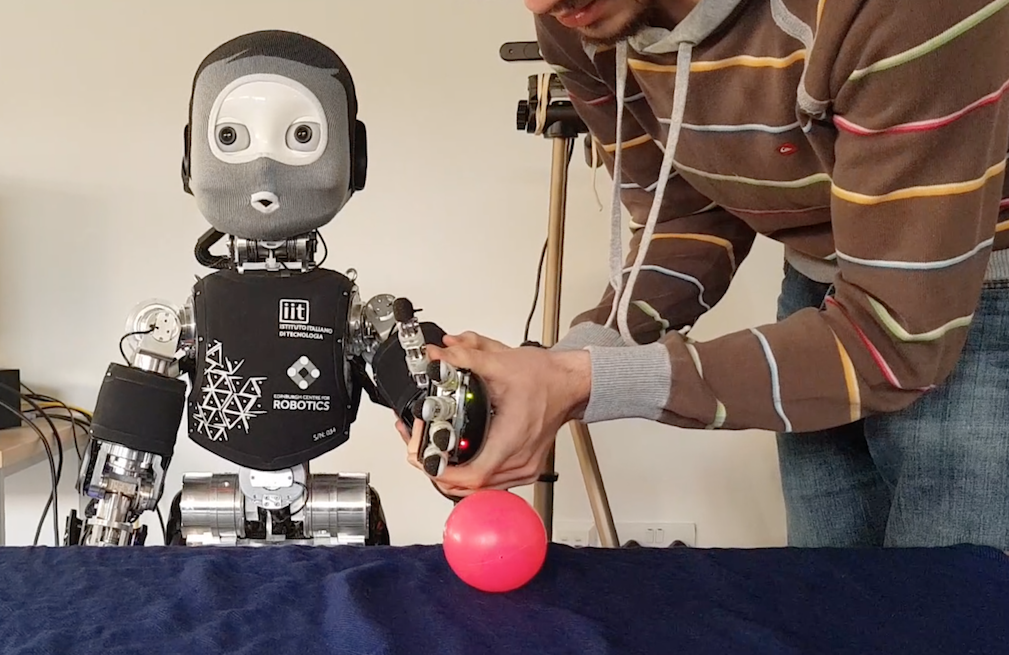}
        \caption{iCub humanoid being taught how to avoid an obstacle (red sphere). Within the proposed framework, this primitive skill provides robustness to novel scenarios.}
        \label{fig:lbd}
    \end{figure}

    The growth of \ac{AI} has popularised more natural techniques for robot learning, reducing the laborious task of coding every possible scenario and thus, increasing modularity and flexibility on the systems. An example of this is imitation learning or \ac{LbD}. This method allows non-robotics-experts to interact, teach and modify the robot’s behaviours~\cite{nicolescu2003natural}, and, consequently, to obtain more human-like behaviours with enhanced acceptability and compatibility to the human workspaces~\cite{ajoudani2017progress}. Given the possibility to learn from humans' expertise and dexterity in using both arms for manipulation purposes, it is natural to exploit \ac{LbD} to use human motions in robotic control.

    Teaching a robot from human demonstrations can be challenging. The different anatomical characteristics between the teacher and the learner produces the correspondence problem, i.e. the issue of identifying a mapping between the teacher and the learner which allows transferring of information from one to the other~\cite{dautenhahn2002correspondence}. Moreover, complex motions involve a mixture of human intentions, which are difficult to accurately learn when following an all-at-once learning baseline~\cite{bajcsy2018learning}. On top of that, teaching a dual-arm system can suppose a high endeavour for non-robotics-experts~\cite{akgun2012keyframe}.

    \ac{LbD} offers some generalisation capabilities, such as changes in initial and goal configurations of a given demonstration~\cite{billard2008robot}. However, being limited to similar scenarios is not realistic to the rapidly changing, dynamic and unpredictable environments where robots have to operate. Extended robustness can be obtained by letting the system iteratively adapt and improve the learnt task to new scenarios~\cite{guenter2007reinforcement}. This leads to the well-known exploration-exploitation dilemma and comes at the cost of needing to fail in order to learn and consequently, at the risk of causing harm to the robot during the self-learning process.

    This paper presents a framework that seeks to jointly overcome the aforementioned issues, namely (i)~the complex and ambiguous teaching procedures and (ii)~the limited generalisation capabilities. Aiming to provide a dual-arm system with a more general and less task-specific method for real-time and robust manipulation in challenging even unfamiliar environments, the proposed framework (i)~leverages human knowledge to learn and create a library of primitive skills and (ii)~endows dual-arm systems with human-like manipulation capabilities by combining the primitive behaviours.

    The main contribution of this work is the formulation of a framework which learns individual primitive skills from human demonstrations and exploits them for robust dual-arm manipulation purposes. Such a framework extends the capabilities of the method in~\cite{pastor2009learning} to handle the requirements of dual-arm systems. This leads to a framework which reuses its knowledge to generalise according to the environment awareness, differently from the proposals in~\cite{zollner2004programming,topp2017knowledge}. The potential of this method has been demonstrated in a simulated environment on the iCub humanoid (see \fref{fig:lbd}). The experimental results suggest the suitability of the framework to address the aforementioned challenges.

	\section{SYSTEM DESCRIPTION {\normalsize AND} PROBLEM FORMULATION} \label{sec:formulation}
    The aim of this paper is an end-to-end learning-based framework that allows real-time autonomous dual-arm manipulation in unfamiliar environments. Such a framework needs to be able to adapt its plan to achieve a task according to the surrounding environment, while ensuring some synchronisation between both end-effectors. Moreover, it needs to be easily programmable, making a dual-arm platform customizable and accessible even to non-robotics-experts. Bearing these problem requirements in mind, this section firstly describes the typology and diversity of possible actions in a dual-arm system. It then analyses the challenges that arise when learning actions from human demonstrations. Finally, this section puts the previous pieces together to formulate the modelisation of the dual-arm system and its grasping.

    \subsection{Dual-arm Primitive Skills Taxonomy}
        Dual-arm manipulators are extremely sophisticated systems, and consequently, their control actions to achieve a specific performance. This work contemplates that any complex behaviour is composed of a vast repertoire of actions or primitive skills~\cite{montesano2008learning}. In the context of manipulation via a dual-arm system, a possible classification of any primitive skill falls into these two groups:
        \begin{itemize}
            \item Absolute skills $\Askill$: imply a change of configuration of the manipulated object in the Cartesian space. Example: move, place and/or turn an object in a particular manner.
            \item Relative skills $\Rskill$: exert an action on the manipulated object in the Object space. Example: opening of a bottle's screw cap, or hold a parcel by means of force contact.
        \end{itemize}

        Each type of primitive skill uniquely produces movement in its own space. In other words, the absolute and relative skills lay on orthogonal spaces. It is natural to expect from a dual-arm system to simultaneously carry out, at least, one absolute and one relative skill to successfully accomplish a task. Let us analyse the task of moving a bottle to a certain position while opening its screw cap. Both end-effectors synchronously move to reach a desired configuration (absolute skill). At the same time, the left end-effector is constrained to hold the bottle upright (relative skill), while the right end-effector unscrews the cap (relative skill).

    \subsection{Learning for a Dual-arm Manipulator}
        \Acf{LbD} provides a large set of recording techniques and mathematical supports for encoding a demonstrated skill. However, learning a particular task from human demonstrations raises some challenges, namely (i)~clearly understanding the intentions of a demonstration and (ii)~establishing a teacher-learner communication channel. Both issues can drastically affect the learning outcome if they are not well adressed~\cite{argall2009survey}.

        The demonstration clarity issue is tackled by leveraging the belief of a vast repertoire of primitive skills being the basis of any complex behaviour. With this in mind, this work avoids demonstrating a task itself but, instead, teaches the robot the involved primitive skills. This task factorisation provides similar benefits as the work in \cite{bajcsy2018learning}: it allows the user to teach one feature of the task at a time, and, if required, to correct them individually.

        Factorising a complex behaviour into primitive actions reduces the number of \ac{DoF} to focus on during demonstration time. As an example, the desired position and orientation of a task can be encoded in separate primitive skills and thus, demonstrated one-at-a-time. This fact becomes handy to ease the complex process of teaching a dual-arm system~\cite{akgun2012keyframe}. This work employs kinesthetic guiding to establish a teacher-learner communication channel which does not suffer from the correspondence problem~\cite{argall2009survey}.

    \subsection{Dual-arm System Modelisation} \label{formulation_model}
		\begin{figure}[t!]
			\centering
			\includegraphics[width=4.5cm]{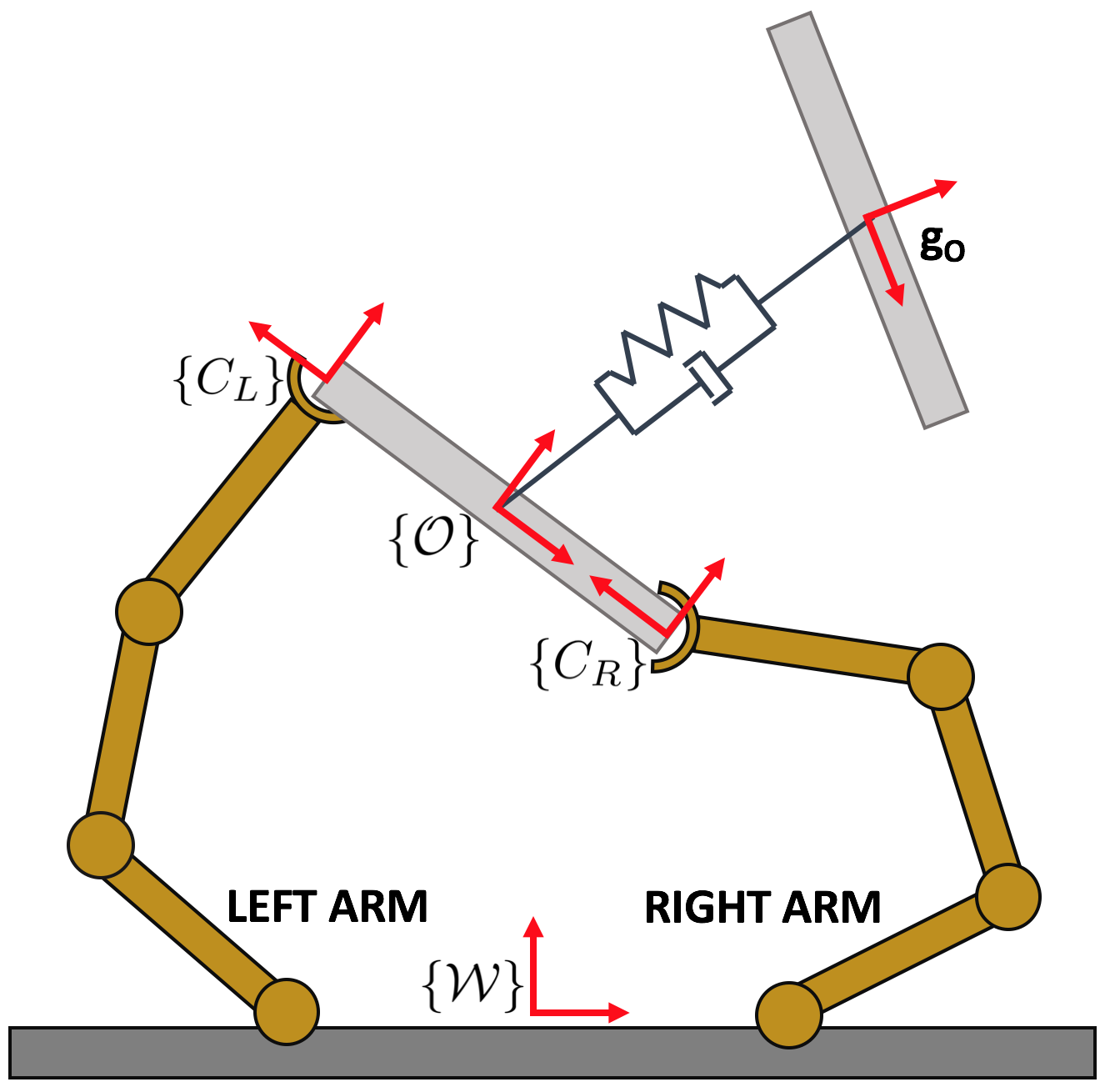}
			\caption{Dual-arm manipulator modelled in the Cartesian space as a spring-damper closed-chain system.}
			\label{fig:system}
		\end{figure}
        Given the variety of primitive skills that a dual-arm system can execute, this work seeks to model the robotic platform in a generalisable yet modular fashion, which accounts for both absolute and relative skills. To this aim, let us consider the closed kinematic chain depicted in \fref{fig:system}. Each arm $i$, where $i=\{L,\;R\}$, interacts with the same object $\object$ in the workspace ${\workspace \in \real^\dimcs}$, where $\dimcs$ is the dimensionality of the considered Cartesian subspace. In this context, the absolute skill explains the movement of the object $\object$ in the workspace $\workspace$, while the relative skill describes the actions of each end-effector $i$ with respect to the object's reference frame $\{\object\}$. Note that $\{\object\}$ is the centre of the closed-chain dual-arm system. Thus, the remaining of the paper uses $\{\object\}$ as object's and system's frame indistinguishably.

        Let the current state of the closed-chain dual-arm system be defined by the position, velocity and accelaration of its system's frame $\{\object\}$ in each \ac{DoF} of the workspace $\workspace$, i.e. ${(\pos_o, \; \vel_o, \; \acc_o)_n \; \forall \; n \in [1, \; \dimcs]}$. The dynamics of such a system are approximated by the ones of a spring-damper system acting between the objects's frame $\{\object\}$ and its goal configuration $\vgoal_o$ (see \fref{fig:system}). This dynamical system genarates in each \ac{DoF} a movement trajectory $\vpos_o$ with velocity $\vvel_o$ and acceleration $\vacc_o$ profiles defined by:
        \begin{align}
            \tau \vacc_o &= \alpha (\beta (g_o - \vpos_o) - \vvel_o), \label{eq:dmp_1}
        \end{align}
        where $g_o$ is the model's attractor that the system will converge to with critically damped dynamics and null velocity when $\alpha > 0$, $\beta > 0$ and $ \beta = \alpha / 4$~\cite{ijspeert2013dynamical}.

        Given any initial system state, the dynamical system in~\eref{eq:dmp_1} generates a linear displacement towards the goal configuration $\vgoal_o$. Any other dynamical behaviour can be represented by an external force acting on the system's frame $\{\object\}$ as:
        \begin{align}
            \tau \vacc_o &= \alpha (\beta (g_o - \vpos_o) - \vvel_o) + \adaptation{_o}, \label{eq:dmp_1_c}
        \end{align}
        where the coupling term $\adaptation{_o}$ describes the profile of the external force affecting the natural dynamics of the system. In other words, $\adaptation{_o}$ characterises the system's behaviour and thus, can be used to encode and retrieve a primitive skill.

    \subsection{Dual-arm Grasping Geometry}
        Any action referenced to the object's frame $\{\object\}$ can be projected to the end-effectors using the grasping geometry of the manipulated object. This allows computing the required end-effector control commands to achieve a particular absolute task. To this aim, the grasp matrix needs to be computed. The grasp matrix $\graspmatrix_i$ of the end-effector $i$ is a transformation map which establishes a velocity relation between the contact point $C_i$, and the systems reference frame $\{\object\}$. For a workspace $\workspace$ of $\dimcs=6$, i.e. considering the linear and rotational information of the \ac{3D} space, the grasping geometry establishes the following relation:
        \begin{align}
            \vvel_{C_i} &= \graspmatrix_i^T \; \vvel_o,
        \end{align}
        where
        \begin{align}
            \graspmatrix_i \in \real^{6 \times 6} =
            \begin{bmatrix}
                \mathbf{I}_{3 \times 3} & \mathbf{O}_{3 \times 3} \\
                \mathbb{S}(\vr_i) & \mathbf{I}_{3 \times 3}
            \end{bmatrix},
        \end{align}
        where $\mathbf{I}_{3 \times 3}$ is the identity matrix, and $\mathbb{S}(\vr_i) \in \real^{3 \times 3}$ is the skew-symmetric matrix performing the cross product:
        \begin{align}
            \mathbb{S}(\vr_i) =
            \begin{bmatrix}
                0 & -r_z & r_y \\
                r_z & 0 & -r_x \\
                -r_y & r_x & 0
            \end{bmatrix},
        \end{align}
        where $\vr_i$ is the distance from the object's reference frame $\{\object\}$ to the contact point $C_i$.

		A global grasp map $\graspmatrix$ for the dual-arm manipulator can be defined by horizontally concatenating the grasp matrix of each end-effector, i.e. ${\graspmatrix = [\graspmatrix_L \; \graspmatrix_R] \in \real^{6 \times 12}}$ where $\graspmatrix_L$ and $\graspmatrix_R$ are the left and right arm grasp matrix, respectively.

	\section{FRAMEWORK FOR ROBUST DUAL-ARM MANIPULATION} \label{sec:learning}
    \begin{figure*}[t]
        \centering
        \includegraphics[width=17.5cm]{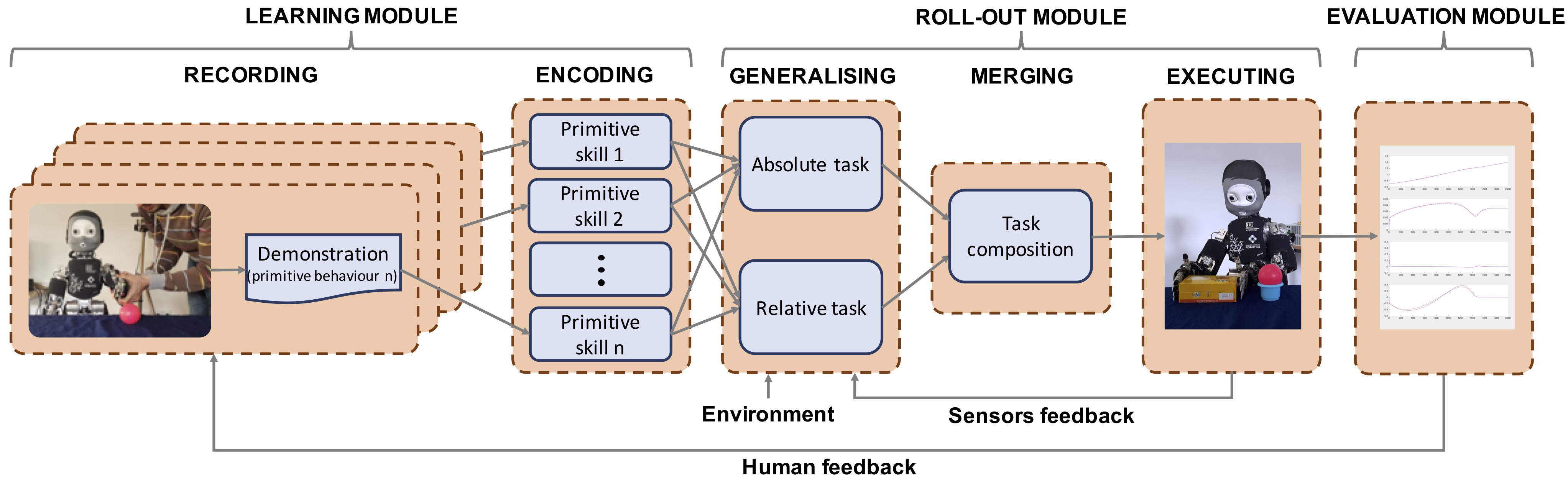}
        \caption{Scheme of the three stages involved in the proposal. Learning: a human demonstrator teaches some primitives behaviours to a system. Rolling-out: the robot exploits (generalises and combines accordingly to the environment awareness) the acquired knowledge. Evaluation: an evaluator inspects the system's performance and decides whether reteaching is necessary.}
        \label{fig:bigpicture}
    \end{figure*}

    \begin{figure}[b!]
        \centering
        \includegraphics[width=7cm]{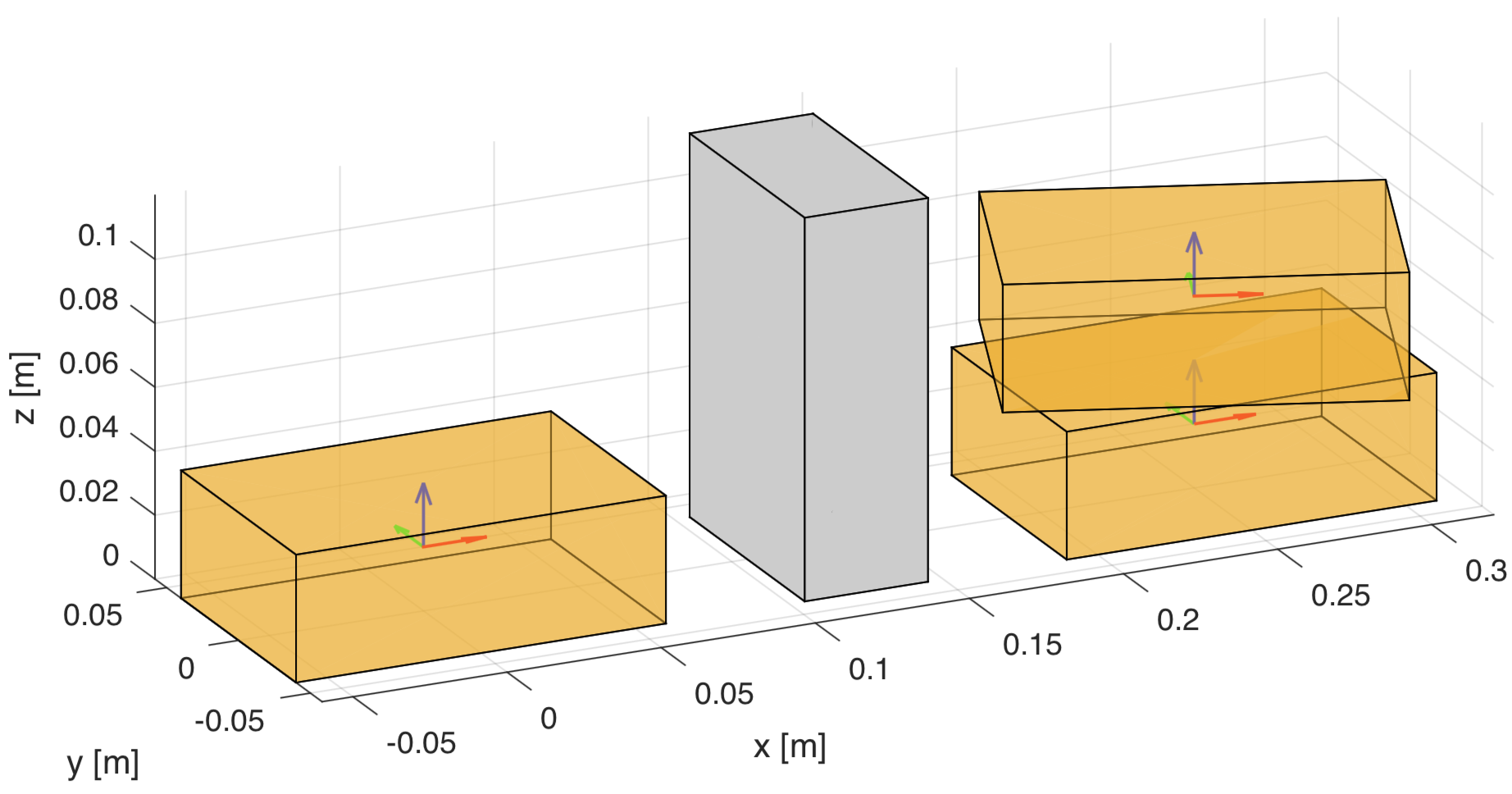}
        \caption{Dual-arm pick-and-place of a parcel (brown prism) in the presence of obstacles (grey prism).}
        \label{fig:main_example}
    \end{figure}

    In order to endow robots with real-time, robust and autonomous dual-arm manipulation, while letting non-robotics-experts to easily program and customise the system's behaviour, this work presents the learning-based framework depicted in~\fref{fig:bigpicture}. Such a framework jointly addresses the aforementioned requirements with three sequential parts: (i)~the learning module that learns a set of primitive skills from human demonstrations, (ii)~the roll-out module that combines those primitive skills to plan a trajectory which makes the system succeed at a task, even in unfamiliar environments and (iii)~the evaluation module that lets a human-in-the-loop supervise the robot's behaviour and reteach a specific skill, if required.

    Given a learnt repertoire (library) of absolute and relative primitive skills, such basic motions need to be combined to confront any dual-arm task in any possible scenario. Each absolute task $\Askill$ is defined by its coupling term $\adaptation{_o}$, which leads to a desired triplet ${(\vacc_o, \; \vvel_o, \; \vpos_o)_n \; \forall \; n \in [1, \; \dimcs]}$ after rolling-out \eref{eq:dmp_1_c}. Similarly, each relative task $\Rskill$ defines a desired triplet for each end-effector $i$ ${(\vacc_{C_i}, \; \vvel_{C_i}, \; \vpos_{C_i})_n \; \forall \; n \in [1, \; \dimcs]}$. This work considers weighting and merging the contribution of each primitive skill at the velocity level as:
    \begin{align}
        \begin{bmatrix}
            \vvel_L \\
            \vvel_R
        \end{bmatrix} =
        \graspmatrix^T \sum_{j=1}^J w_j \; \vvel_{o_j} + \sum_{k=1}^K w_k
        \begin{bmatrix}
            \vvel_{C_{L,k}} \\
            \vvel_{C_{R,k}}
        \end{bmatrix},
        \label{eq:merge}
    \end{align}
    where $\vvel_L$ and $\vvel_R$ respectively are the velocity commands for the left and right end-effector which satisfy the set of activated primitive skills, $\vvel_{o_j}$ is the velocity of the $j \in [1, \; J]$ absolute primitive skills available in the library, and $\vvel_{C_i,k}$ is the velocity of the $k \in [1, \; K]$ relative primitive skills available in the library. Primitive skill selection according to a desired task and environment is conducted with the weights $w_j$ and $w_k$. Works such as the one in~\cite{ardon2018towards} propose addressing this problem according to the object's affordances and environment analysis.

    The generality of the proposed framework is narrowed down to provide an application case. This work exploits such a framework to endow a dual-arm system with enhanced autonomy for the dual-arm task of pick-and-place of a parcel, even in the presence of unexpected obstacles. \fref{fig:main_example} depicts the main idea: parcels (brown prisms) are meant to be moved from one side to another, adjusting the behaviour of the dual-arm whether there is an obstacle or not (grey prism). Not requiring complex grasping capabilities is the main reason for choosing this application case. However, it is extremely challenging in the synchronisation aspect: manipulators have to always maintain a certain amount of contact forces with the carried parcel as any variation would result in releasing or exposing the handled object to stress. To this aim, the library of primitive skills is loaded with: underlying dynamics of a pick-and-place task, obstacle avoidance and grasp maintenance behaviours. Note that the former two skills are absolute, while the latter is relative.

    \subsection{Skill Dynamics}
\begin{figure}[t!]
			\centering
			\includegraphics[width=7.5cm]{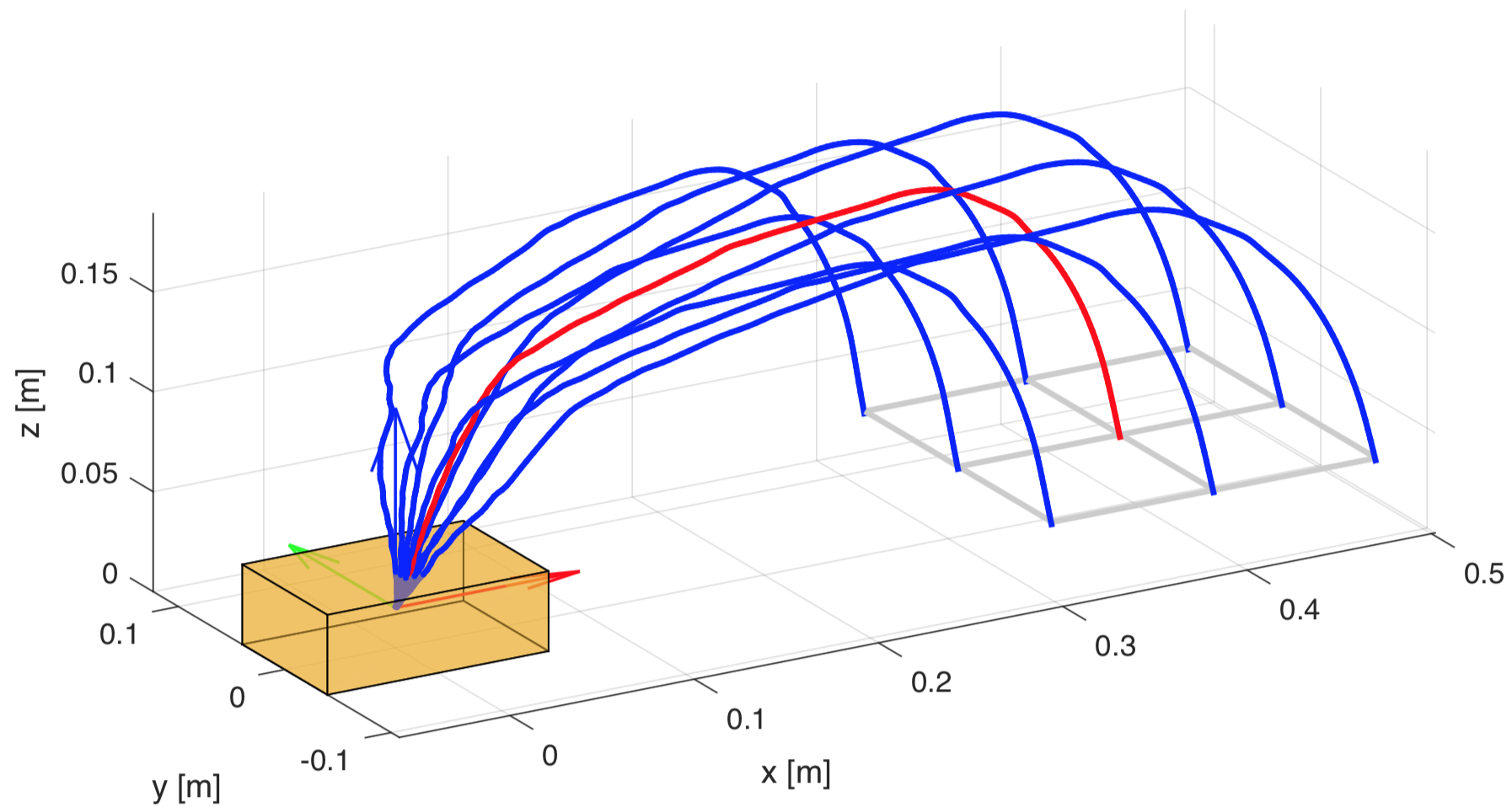}
			\caption{\ac{DMP} generalisation capabilities. Given a demonstration (red trajectory), rolling-out \eref{eq:dmp_1_c} with the \ac{DMP} coupling term $\adaptation{_o}$ defined in \eref{eq:dmp_rbf} let the system generalise to new goal configurations (blue trajectories).}
			\label{fig:dmp_generalisation}
		\end{figure}

        The non-linear dynamical behaviour of any task can be represented using \acp{DMP}. This mathematical encoding support has proven to be a versatile tool for modelling and learning complex motions, since: (a)~any movement can be efficiently learned and generated, (b)~a unique demonstration is already generalisable, (c)~convergence to the goal is guaranteed, and (d)~their representation is translation and time-invariant~\cite{pastor2009learning,ijspeert2013dynamical}. Some of these \ac{DMP}-inherent generalisation capabilities are depicted in \fref{fig:dmp_generalisation}.

		The system modelisation in~\eref{eq:dmp_1_c} can integrate a \ac{DMP} as the coupling term $\adaptation{_o}$. This means that the perturbationless dynamics of the spring-damper system are modified according to the \ac{DMP} coupled in each \ac{DoF}. If ${\workspace \in \real^3}$, three position-encoding \acp{DMP} would describe the desired position of the manipulated object. Instead, if ${\workspace \in \real^6}$, four additional quaternion-based \acp{DMP} would be required to also encode the object's desired orientation~\cite{ude2014orientation}.

        Formally, a position-encoding \ac{DMP} is a weighted linear combination of non-linear \acp{RBF}~\cite{pastor2009learning,ijspeert2013dynamical}. The value of such non-linear function $\adaptation{_o}$ when evaluated at a specific entry ${k \in \vk}$ is defined as:
        \begin{align}
            f(k) &= \frac{\sum_{i=1}^N w_i\Psi_i(k)}{\sum^N_{i=1} \Psi_i(k)} \; k, \label{eq:dmp_rbf} \\
            \Psi_i(k) &= \exp\mathopen{}\left(-h_i(k-c_i)^2 \right)\mathclose{},
        \end{align}
        where $c_i$ and $h_i > 0$ are the centres and widths, respectively, of the $i \in [1, \; N]$ \acp{RBF} distributed along the trajectory. Each \ac{RBF} is weighted by $w_i$. The phase variable $\vk$ is utilised to avoid direct dependency of $\adaptation{_o} \sim \vf(\vk)$ on time. The dynamics of $\vk$ are defined as:
        \begin{align}
            \tau \dot{\vk} = -\alpha_k \vk,
        \label{eq:canonicalsystem}
        \end{align}
        where the initial value of the canonical system $\vk(0) = 1$ and $\alpha_k$ is a positive constant.

		The learning of the \acp{DMP} relies on adjusting the set of \ac{RBF}, i.e. the weight vector $\vw$, composed of all weights $w_i$. To this aim, \ac{LMS} is used to compute the weight vector $\vw$ which makes the system~\eref{eq:dmp_1_c} adjust to a recorded skill propioception information $\{\vacc, \; \vvel, \; \vpos\}$.

    \subsection{Obstacle Avoidance}
		\begin{figure}[b!]
			\centering
			\includegraphics[width=4cm]{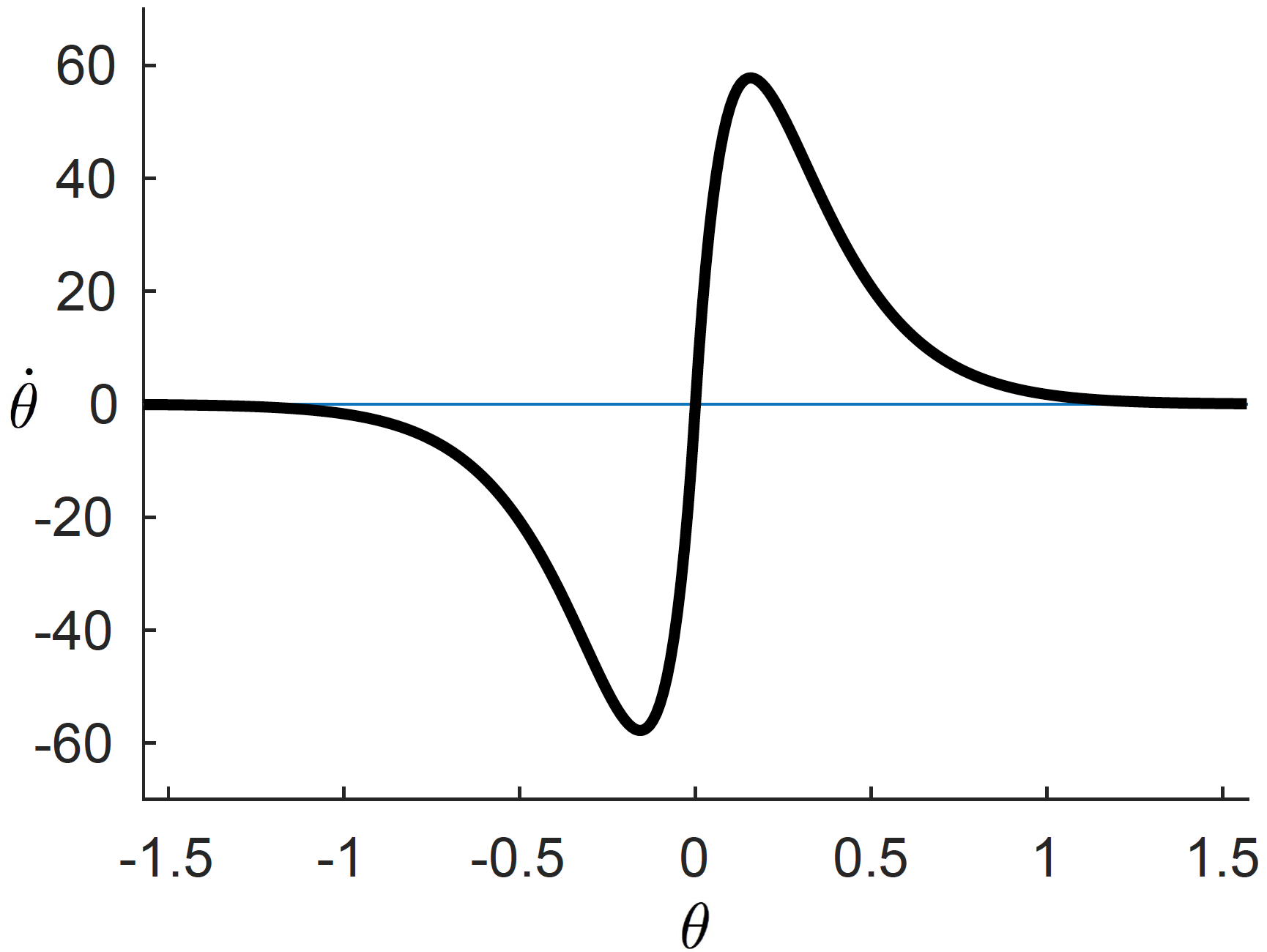}
			\caption{Change of steering angle $\dot{\theta}$ following the original formulation in~\eref{eq:oa_original} with ${\gamma=1000}$ and ${\beta=20/\pi}$.}
			\label{fig:oa_ori}
		\end{figure}
        An analytical description of how humans steer around an obstacle was first presented in~\cite{fajen2003behavioral}. Later on, such biologically-inspired formulation was used in~\cite{hoffmann2009biologically} for single-arm manipulation purposes. Let $\vpos$, $\vvel$, and $\theta$ be respectively the system's position, velocity and orientation referenced to the workspace reference frame $\{\workspace\}$. In order to avoid an obstacle, the system in~\eref{eq:dmp_1_c} needs to change its acceleration accordingly to:
          \begin{align}
               \vf(\vpos, \; \vvel) = \mathbf{R} \; \vvel \; \dot{\theta}, \label{eq:oa_main}
          \end{align}
          where $\mathbf{R}$ is a $\pi/2$ rotation matrix with respect to the vector $\mathbf{r} = (\vpos_{obstacle} - \vpos) \times \vvel$, and $\dot{\theta}$ is the desired turning velocity:
          \begin{align}
              \dot{\theta} = \gamma \; \theta \exp \!\left(-\beta \; |\theta| \right)\!, \label{eq:oa_original}
          \end{align}
          where $\gamma$ and $\beta$ are tuning constants. Their effect can be best understood in \fref{fig:oa_ori}: $\gamma$ sets the abruptness of the obstacle avoidance behaviour, and $\beta$ determines its sensitivity.

		  Within the framework, the parameters of the obstacle avoidance behaviour are leant from human demonstrations, thus involving less parameter tuning. This is achieved using \ac{LMS} after log-linearising \eref{eq:oa_original} and arranging it as:
        \begin{align}
            \log \dot{\bm{\theta}} =
            \begin{bmatrix}
                \log \gamma & 1 & \beta
            \end{bmatrix}
			\begingroup
			\renewcommand*{\arraystretch}{1.3}
            \begin{bmatrix}
                \mathbf{1} \\
                \log \bm{\theta} \\
                -\bm{|\theta|}
            \end{bmatrix}\!\!,
			\endgroup
         \label{eq:oa_complete_learning}
        \end{align}
        where the training data $\dot{\bm{\theta}}$ and $\bm{\theta}$ contain the periodically sampled value of $\dot{\theta}$ and $\theta$ experienced during the obstacle avoidance demonstration. The change in steering angle $\dot{\theta}$ is retrieved from \eref{eq:oa_main}, where ${\vf(\vpos, \; \vvel) = \vf(\vx)_{obs} - \vf(\vx)}$, i.e. the difference on the dynamics between a perturbationless task $\vf(\vx)$ and one with obstacles $\vf(\vx)_{obs}$ is only motivated by the presence of an obstacle.

    \subsection{Grasp Maintenance}
        Manipulation of a rigid object via a dual-arm system requires each end-effector to be in contact with the object. Moreover, when the interaction is by force contact (without grasping the object) it is also essential to apply the sufficient forces to ensure grasp maintenance, i.e. prevention of contact separation and unwanted contact sliding. The complexity of this task usually requires modelling the required coupling forces as $\vf(\vpos, \; \vvel, \; \vacc)$. For applications with low-dynamical requirements, the previous dynamical function can be approximated to~\cite{gams2014coupling}:
        \begin{align}
            \vvel_{C_{i}} = \mathbf{K} (\mathbf{F}_d - \mathbf{F}_r), \label{eq:grasp_m}
        \end{align}
        where $\mathbf{K}$ is an error multiplying constant which transforms errors in force contact to velocity commands, $\mathbf{F}_d$ is the desired coupling force and $\mathbf{F}_r$ is the current coupling force retrieved from the robot's sensors. Thus, the learning of this primitive skill resides on learning from demonstrations which $\mathbf{F}_d$ ensures grasp maintenance.

	\section{RESULTS {\normalsize AND} EVALUATION} \label{sec:results}
    The work presented in this paper is a generic framework for any dual-arm manipulator. Experimental evaluation has been carried out on synthetic environments and a simulated iCub humanoid. This section firstly introduces the iCub robot and the execution of kinesthetic learning on this platform. It then describes the learning of the obstacle avoidance behaviour, and it analyses its integration in a synthetic pick-and-place task. Finally, this section depicts the potential of the proposed framework for being used on a humanoid robot.

    \subsection{Experimental Platform}
        iCub is an open source humanoid robot testbed for research into human cognition and artificial intelligence applications~\cite{metta2008icub}. The physical and software characteristics of this robot make it an ideal platform for the presented research. Among all this robot's features, some of the most relevant to this work are the two 7-\ac{DoF} manipulators equipped with a torque sensor on the shoulder, tactile sensors in the fingertips and palm, and integrated stereo vision. iCub operates under the \acs{YARP} middleware.

        Kinesthetic teaching on the iCub humanoid is conducted by setting all joints in gravity compensation allowing the teacher to physically manoeuvre the robot through a desired skill. During the demonstrations, proprioception information is retrieved via \ac{YARP} ports.

    \subsection{Obstacle Avoidance Behaviour}
		\begin{figure}[t!]
			\centering
			\subfigure[]{\includegraphics[width=4cm]{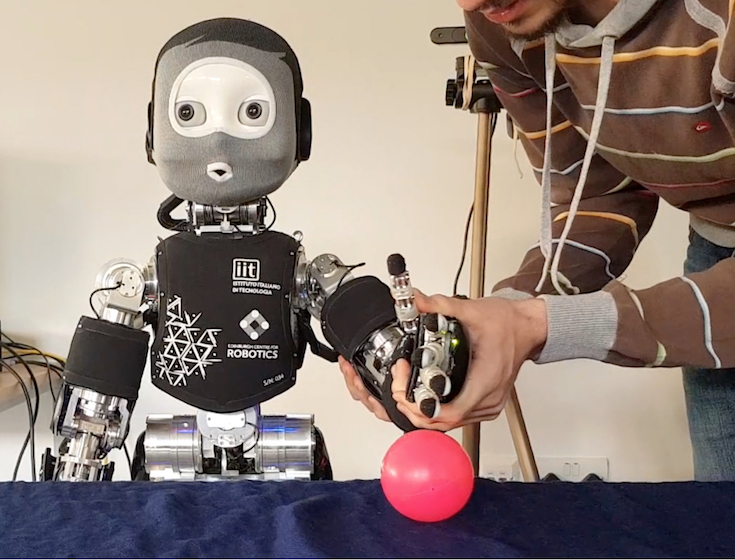} \label{fig:oa_r}}
			\,
			\subfigure[]{\includegraphics[width=4cm]{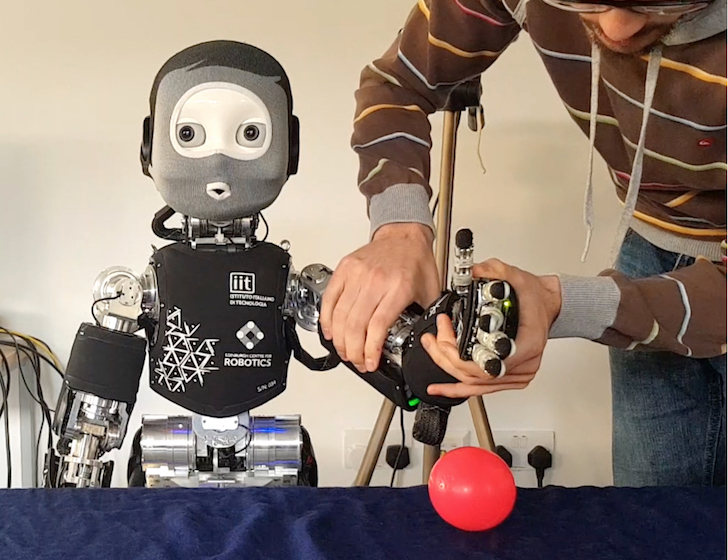} \label{fig:oa_c}}
			\\
			\subfigure[]{\includegraphics[width=4cm]{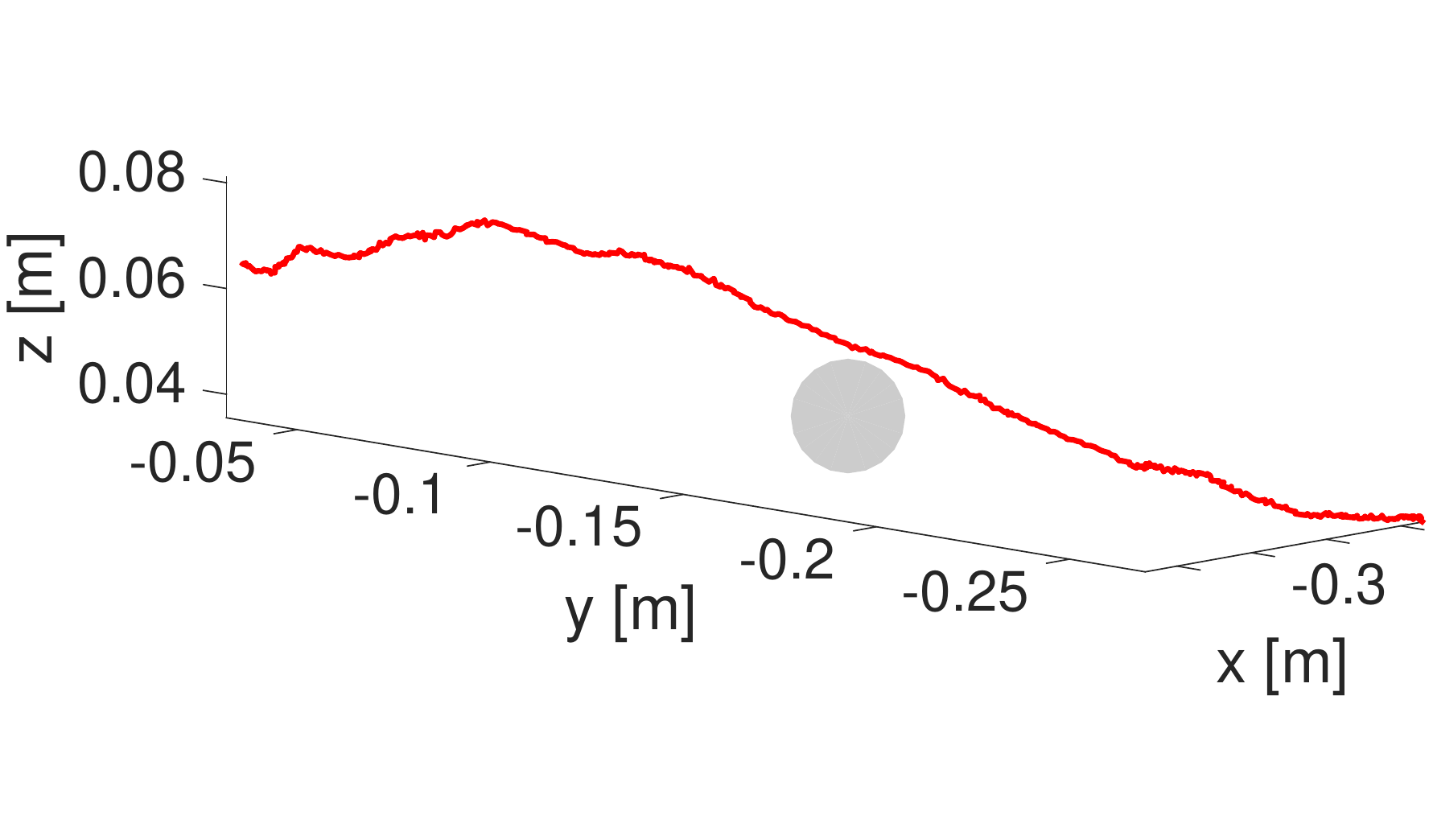} \label{fig:oa_r_raw}}
			\,
			\subfigure[]{\includegraphics[width=4cm]{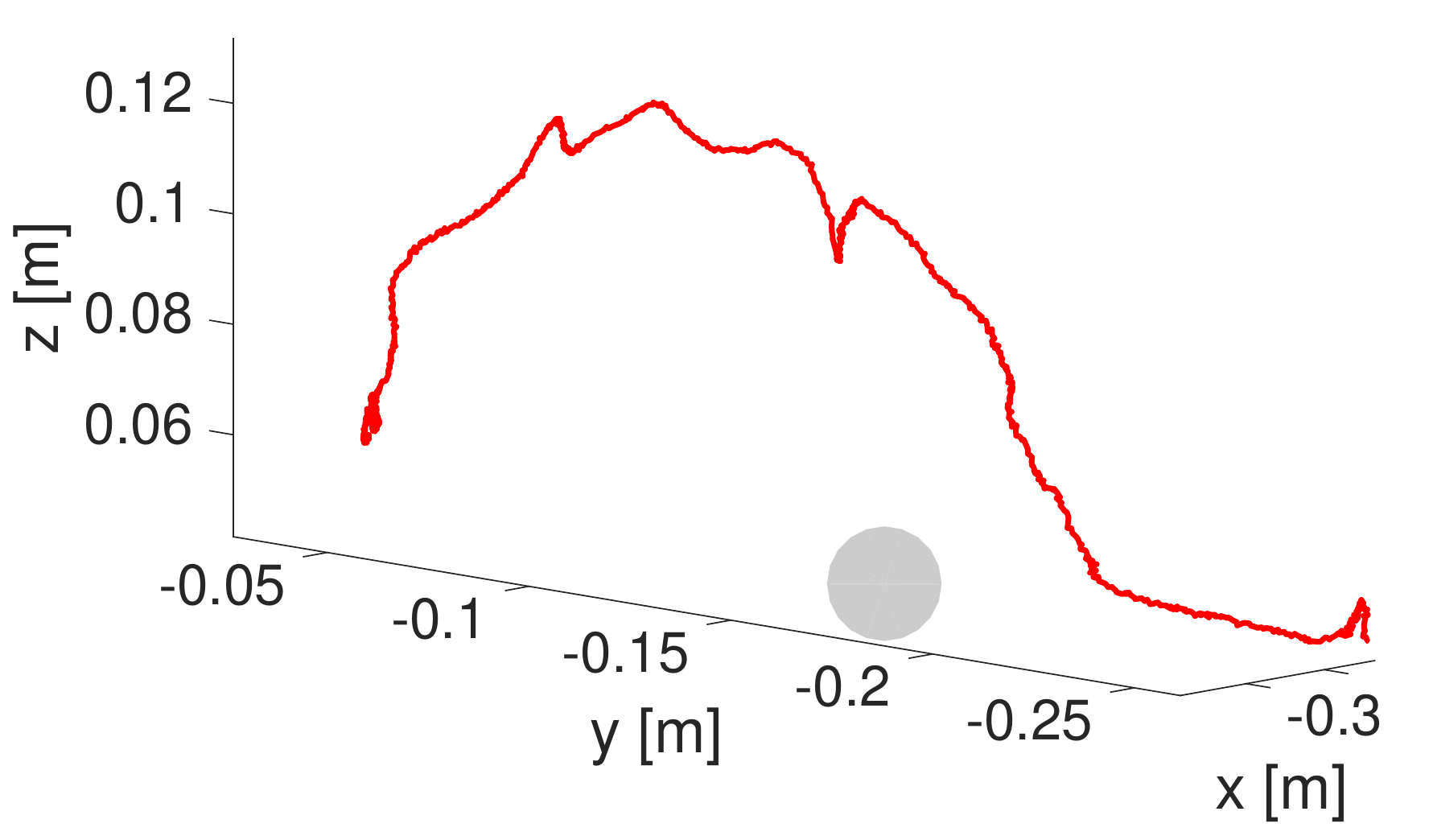} \label{fig:oa_c_raw}}
			\\
			\subfigure[]{\includegraphics[width=4cm]{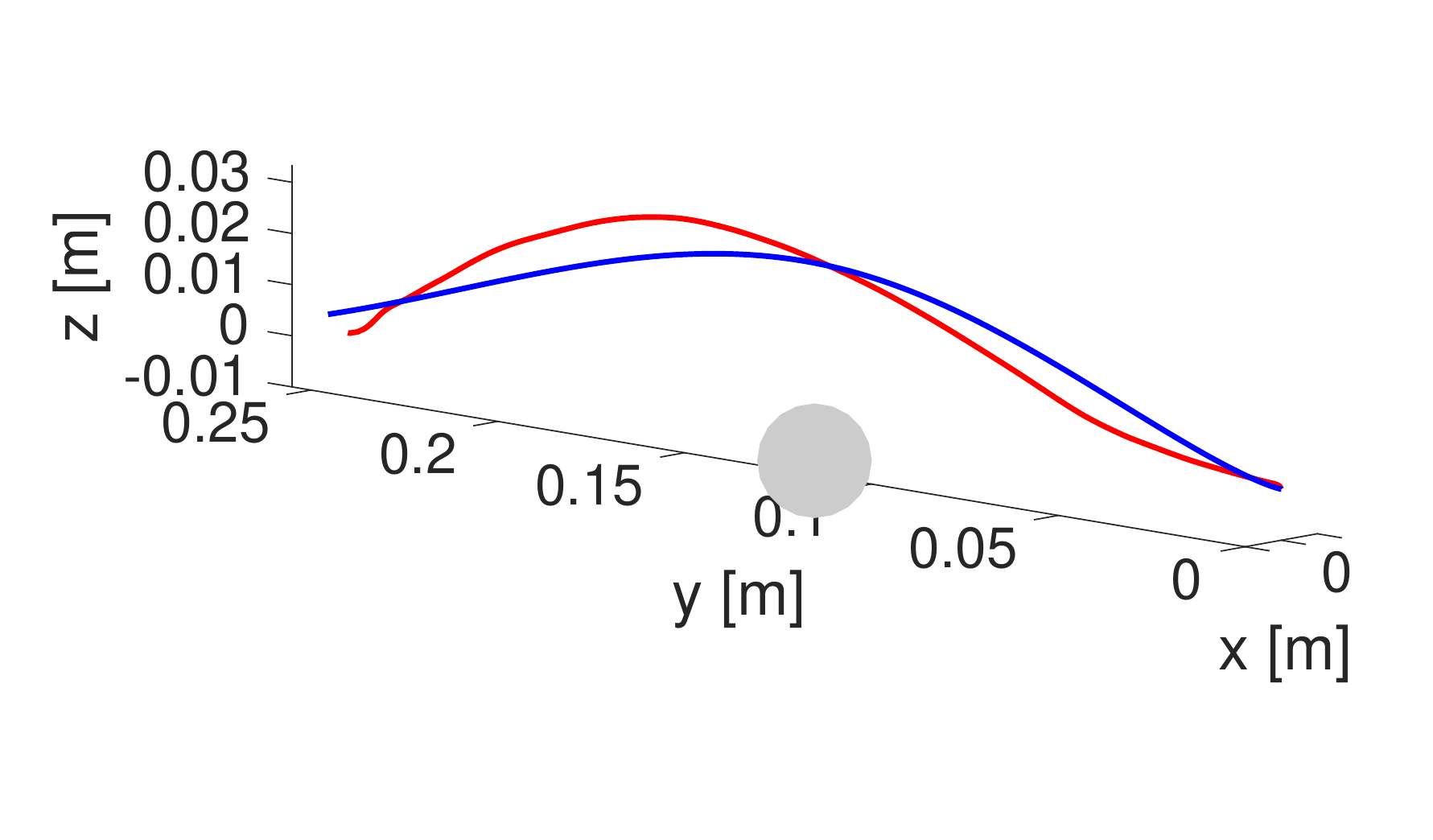} \label{fig:oa_r_ler}}
			\,
			\subfigure[]{\includegraphics[width=4cm]{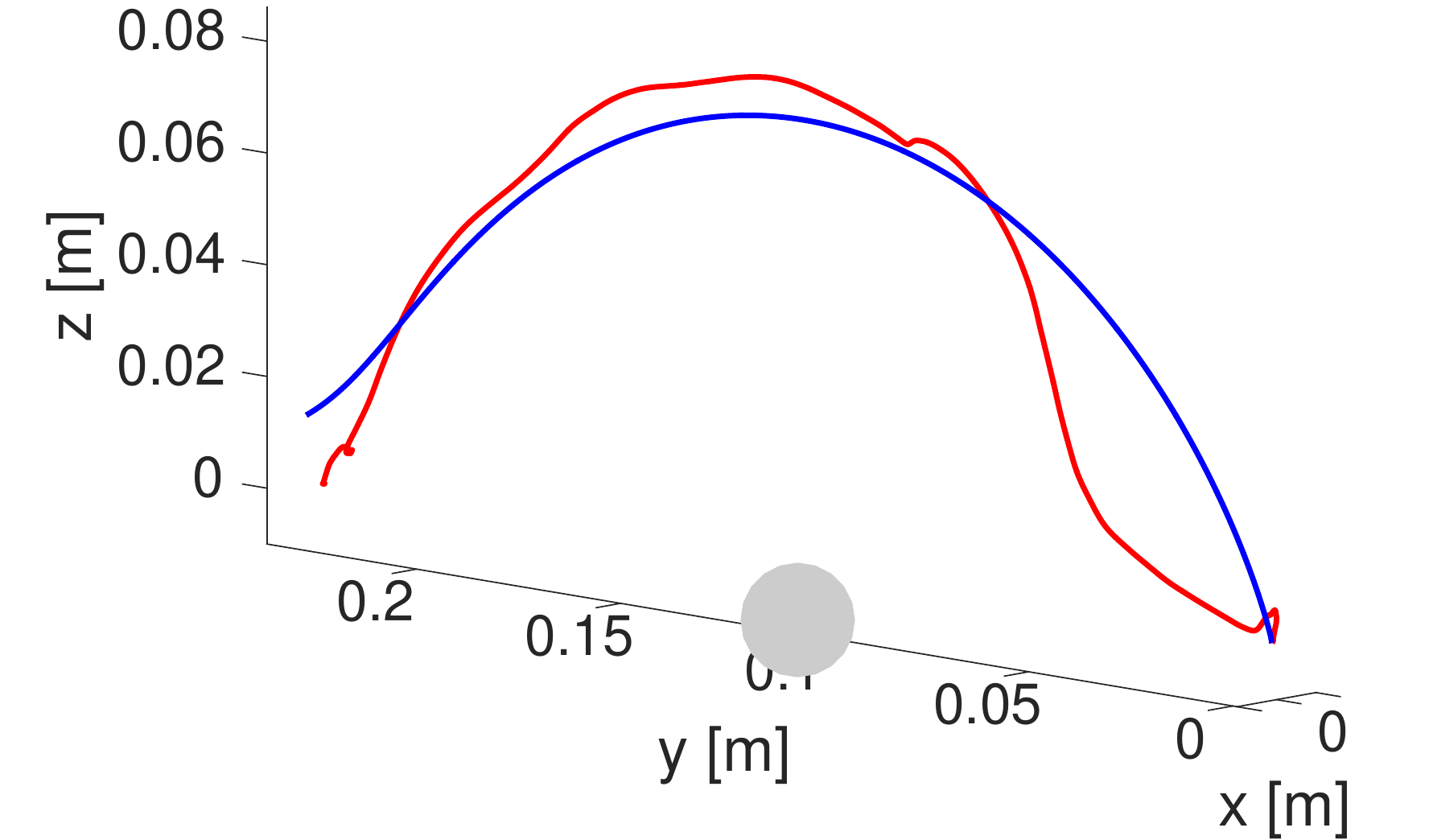} \label{fig:oa_c_ler}}
			\caption{iCub humanoid robot~\cite{metta2008icub} learning the primitive skill of obstacle avoidance with two different behaviours: reckless (first column) and convervative (second column). \mbox{(a)-(b)}~Human demonstrations to avoid an obstacle (red sphere). \mbox{(c)-(d)}~iCub's proprioception data. \mbox{(e)-(f)}~Processed iCub's proprioception data (red trajectory) and learned obstacle avoidance behaviour (blue trajectory).}
			\label{fig:icub_oa_data}
		\end{figure}

        The primitive skill of obstacle avoidance has been taught to iCub with two different behaviours: reckless (see \fref{fig:oa_r}) and conservative (see \fref{fig:oa_c}). While the former steers around the obstacle (red sphere) closely, the latter keeps a larger distance to it. The recorded raw proprioception data of these two kinesthetic demonstrations is respectively portraited in \fref{fig:oa_r_raw} and \fref{fig:oa_c_raw}. As it can be observed, the retrieved trajectories are noisy and not smooth.

        To learn from these demonstrations, the data has been preprocessed in two steps: (i)~filtering to remove outliers and high-frequency noise, and (ii)~projecting the filtered information to the \ac{2D} space composed for the two principal components of the data. \fref{fig:oa_r_ler} and \fref{fig:oa_c_ler} show the preprocessed data (red trajectory), which has been used in~\eref{eq:oa_complete_learning} to learn the parameters defining the demonstrator's obstacle avoidance behaviour. The encoded reckless and conservative styles are respectively depicted in \fref{fig:oa_r_ler} and \fref{fig:oa_c_ler} (blue trajectory). Note that learning the parameters instead of the motion itself lets the robot generalise such behaviour under different conditions.

        In overall terms, from \fref{fig:icub_oa_data} it can be concluded that the obstacle avoidance encoding support and its learning process from human demonstrations is able to encapsulate the demonstrator style. The differences between the demonstrated skill and the learnt one are mainly attributed to the hypothesis that any steering around an obstacle follows the formulation in~\eref{eq:oa_main}-\eref{eq:oa_original}. Moreover, the noise in the proprioception data increases the variance in the learning. Alternatively, a high-precision tracking system such as the one used in~\cite{rai2014learning} can be considered. Because the proposed approach extracts the parameters of an obstacle avoidance behaviour, the resulting knowledge would yet be independent of the demonstration frame.

    \subsection{Synthetic Pick-and-Place Task}
\begin{figure}[h!]
			\centering
			\includegraphics[width=7cm]{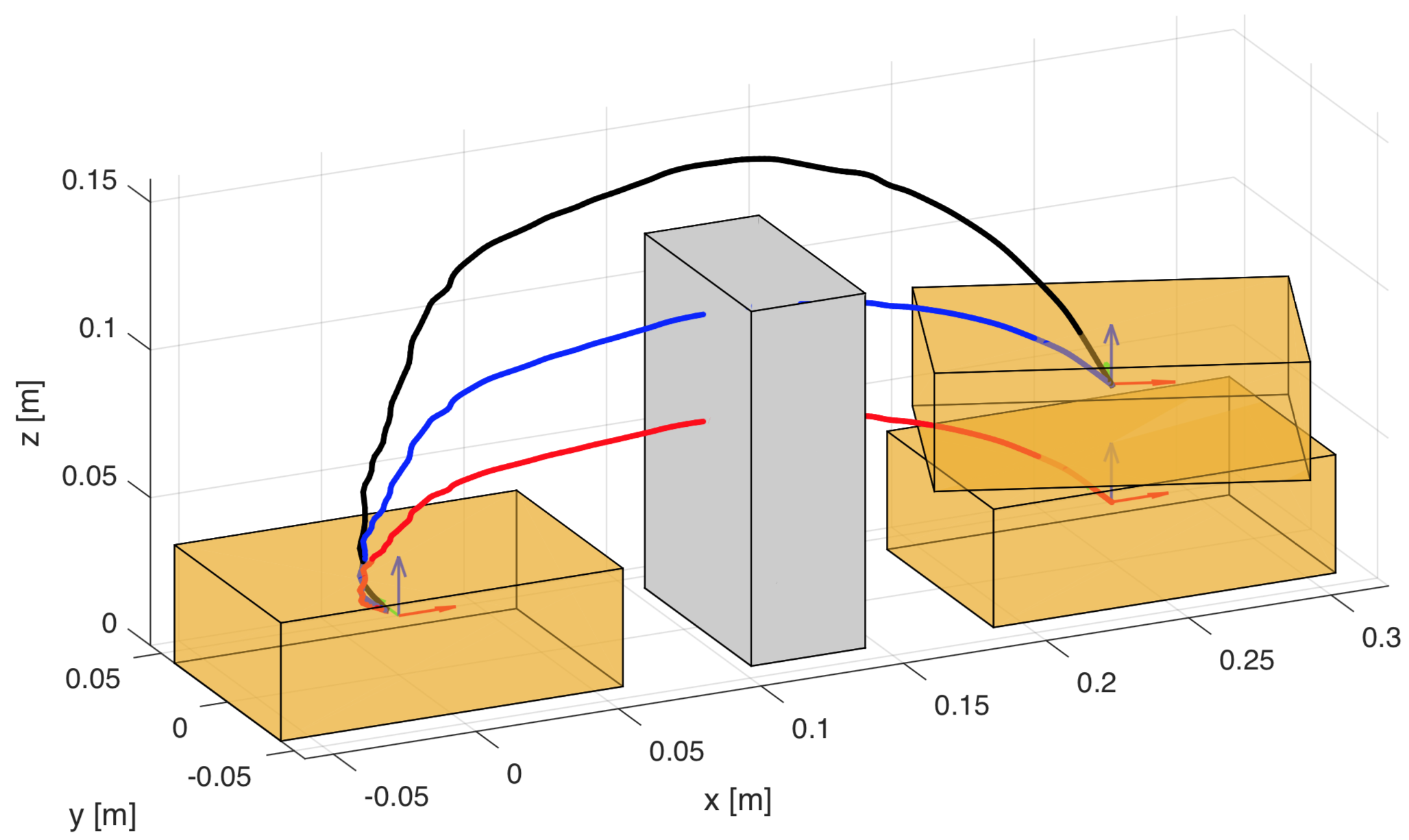}
			\caption{Dual-arm pick-and-place of a parcel (brown prism) in the presence of obstacles (grey prism). Demonstration (red trajectory), inference to a new position (blue trajectory), inference with obstacle avoidance (black trajectory).}
			\label{fig:main_result}
		\end{figure}

        The performance of the obstacle avoidance behaviour in a more realistic context has been validated using the pick-and-place setup depicted in \fref{fig:main_result}. Particularly, an initial pick-and-place demonstration is given to the system (red trajectory), consisting of moving the parcel from the left to the right without the presence of any obstacle (grey prism). The underlying dynamics defining this primitive skill have been encoded as a \ac{DMP}. Due to the inherent generalisation capabilities of the \acp{DMP}, the system is already able to infer the pick-and-place dynamics from any different starting and goal positions (blue trajectory), but not able to generalise to the presence of obstacles. Only after coupling the previously learnt pick-and-place dynamics and obstacle avoidance behaviour together, the system is able to generalise in real-time to the presence of unexpected obstacles (black trajectory).

    \subsection{Framework on Humanoid Robot}
        \begin{figure}[t!]
            \centering
            \subfigure[]{\includegraphics[width=3.5cm]{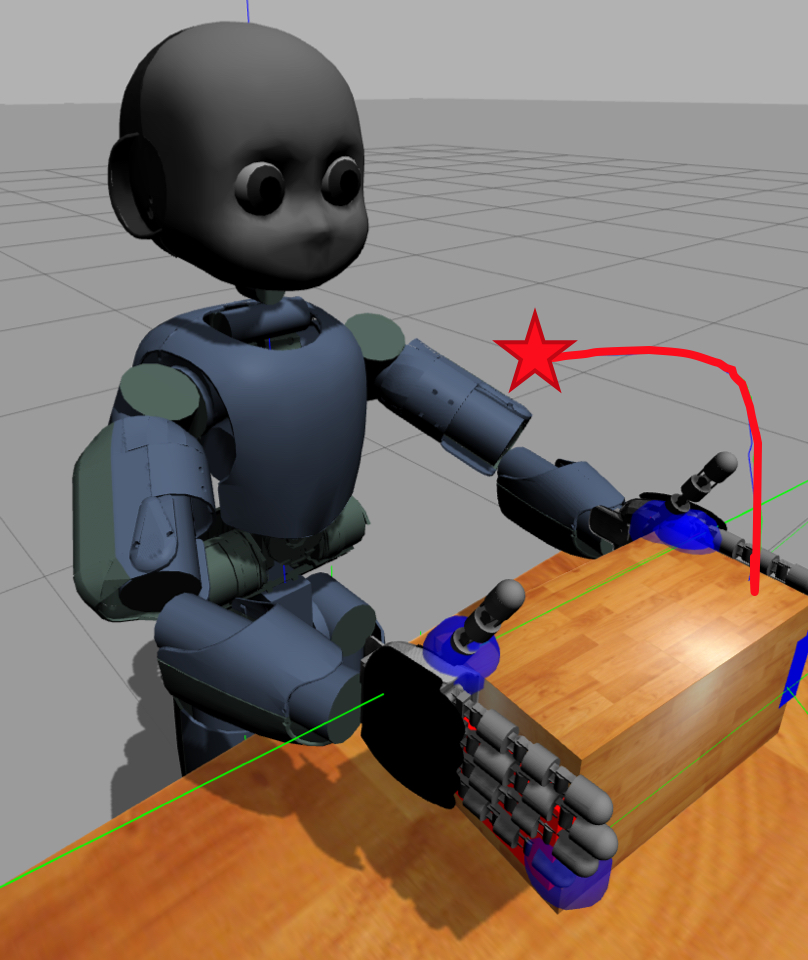} \label{fig:icub_results_a}}
            \quad
            \subfigure[]{\includegraphics[width=3.5cm]{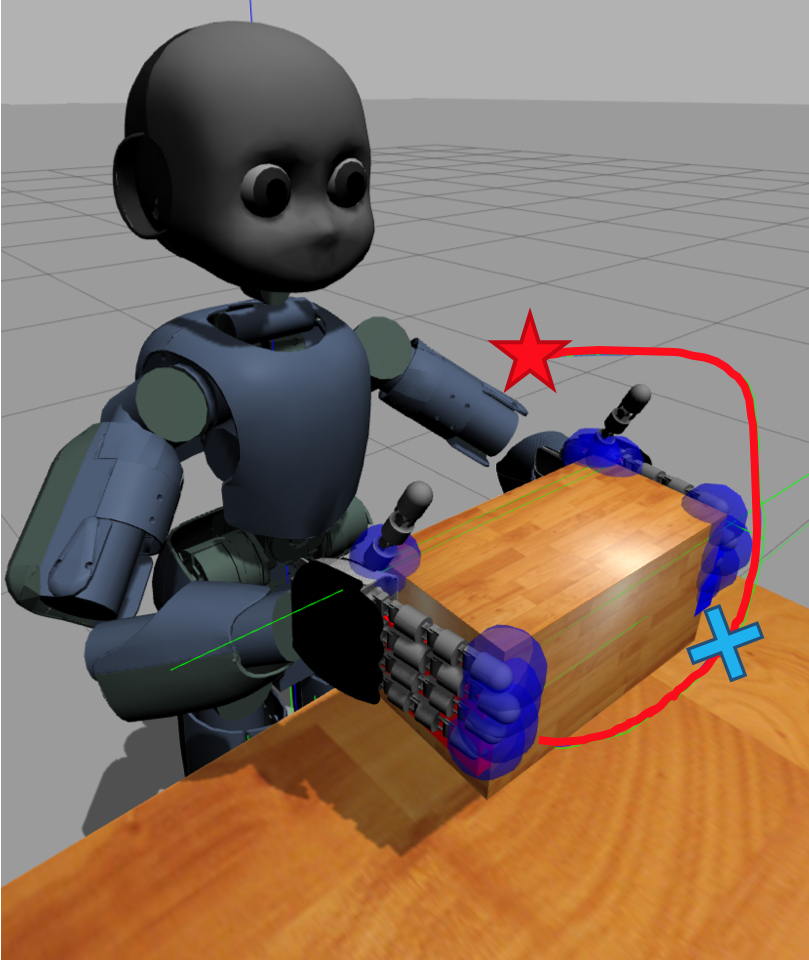} \label{fig:icub_results_b}}
            \caption{iCub humanoid robot~\cite{metta2008icub} picking a parcel and raising it with specific dynamics (red trajectory) to a goal configuration (red star). (a)~Following the task dynamics previously learnt from a human demonstrator. (b)~Modifying the task dynamics in real-time to avoid an obstacle (blue cross). Grasp maintenance is successfuly ensured in both cases by the corresponding primitive skill.}
            \label{fig:icub_results}
        \end{figure}

        The applicability of the framework has been tested with a particular dual-arm task. The framework has been developed in \ac{YARP} to deploy it on a simulated iCub humanoid. Due to the lack of realistic simulated force sensors and thus, lack of awareness of the exerted force on the carried object, the grasp maintenance skill primitive in \eqref{eq:grasp_m} has been replaced according to the proposal in~\cite{gams2014coupling}:
        \begin{align}
            \vvel_{C_{i}} = \mathbf{K} (\mathbf{D}_{d_i} - \mathbf{D}_{r_i}), \label{eq:grasp_m_d}
        \end{align}
        where $\mathbf{D}_{d_i}$ is the desired distance from the object's frame $\{O\}$ to the contact point $C_{i}$, being $i=\{L, \; R\}$, and $\mathbf{D}_{r_i}$ is the current distance. Due to the symmetry of the task, ${\mathbf{D}_{d_L} = \mathbf{D}_{d_R}}$. Thus, the learning of this primitive skill is reduced at setting $\mathbf{D}_{d_i}$ accordingly to the characteristics of the manipulated parcel and the grasping points.

        After this arrangement forced by the simulated nature of the experimentation, the pick-and-raise activity was conducted (see \fref{fig:icub_results}). Such a task consists of picking a parcel from the table and raising it with certain dynamics (red trajectory), while avoiding obstacles and ensuring grasp maintenance. iCub performed the described dual-arm task in two different contexts. First, with the absence of obstacles, where the robot can move the parcel with the designated dynamics (see \fref{fig:icub_results_a}). Second, with the presence of unexpected obstacles (blue cross), where the robot had to replan the trajectory to achieve the goal configuration (see \fref{fig:icub_results_b}). Despite the simplicity of the used primitive skill to ensure grasp maintenance, the trials were successful: both end-effectors were accurately synchronised so the handled parcel was neither released nor exposed to stress.

		    These results show that iCub has been able to perform the pick-and-place task even in the presence of an unexpected obstacle, after learning three primitive skills individually from a human demonstrator. This fact raises expectations about the degree of similarity that iCub's final behaviour might have with the demonstrator's behaviour under the same conditions. Analysing this similarity is of interest to the \ac{HRI} community, since it can contribute to enhancing the acceptability and compatibility of robots in human workspaces~\cite{ajoudani2017progress}. An alternative for conducting this study consists of recording some samples of the robotic and human approach to quantify their deviation with the \ac{KL} divergence statistic. The lower this indicator is, the higher the chances are that these two agents have similar behaviours. Such a study is left for future work.

	\section{FINAL REMARKS {\normalsize AND} FUTURE WORK} \label{sec:final_remarks}
    This work has presented a novel framework which endows a dual-arm system with real-time, robust and less task-specific manipulation capabilities. Such a framework is twofold: (i)~learns from human demonstrations to create a library of primitive skills, and (ii)~combines such knowledge to confront challenging unfamiliar scenarios with human-like manipulation capabilities. Unlike the framework of motion primitives in~\cite{pastor2009learning}, the proposed approach handles primitive skills for dual-arm manipulation purposes while still being able to combine different primitives at the same time. This feature is what differentiates the current work from similar ones~\cite{zollner2004programming,topp2017knowledge}. The evaluation conducted on the iCub humanoid suggest the proposal's suitability for robust dual-arm manipulation, yet with some room for improvement.

    The framework is not restricted to the presented experimental evaluation nor platform. Any system able to retrieve proprioception information can benefit from this work. Moreover, any primitive skill that might be required for dual-arm manipulation can be included in the framework's library. The application case reported in this manuscript exemplifies this fact by considering, among other primitive skills, an obstacle avoidance behaviour which steers around obstacles in real-time. The desired reactivity of this obstacle avoidance behaviour is learnt from human demonstrations.

    Future work will significantly extend the library of primitive skills such that more tasks and scenarios involving challenging dual-arm manipulation tasks can be addressed within the framework. Action selection will be integrated to automatically select from the framework's library the necessary set of skills to address a specific task. Apart from the task itself, surrounding environment and characteristics and constraints of the object to manipulate might need to be considered. Finally, imminent efforts will focus on learning force-dependant primitive skills, such as the grasp maintenance one, on the real iCub humanoid robot, as well as evaluating the entire framework on such platform.

	\section{ACKNOWLEDGMENTS}
	This work has been partially supported by ORCA Hub \mbox{EPSRC} (EP/R026173/1) and consortium partners.

	\bibliographystyle{Format/aaai}
	\bibliography{refs}
\end{document}